\documentclass{article}
\usepackage{spconf,amsmath,graphicx,svg,xprintlen,tabularx}
\usepackage[pagebackref,breaklinks,colorlinks]{hyperref}

\usepackage{tikz,tikzpagenodes,dblfloatfix,multirow}
\usetikzlibrary{calc,positioning,shapes,decorations.pathreplacing}
\definecolor{myblue}{RGB}{71, 181, 255}
\definecolor{mylightblue}{RGB}{223, 246, 255}
\definecolor{myred}{RGB}{234,51,35}
\definecolor{myotherred}{RGB}{234,51,100}
\definecolor{myyellow}{RGB}{245, 194, 66}

\DeclareMathOperator{\lcm}{lcm}

\title{Multistage Spatial Context Models for Learned Image Compression}
\name{Fangzheng Lin\textsuperscript{1}, Heming Sun\textsuperscript{2,3}, Jinming Liu\textsuperscript{1}, Jiro Katto\textsuperscript{1,2}\thanks{Heming Sun is the corresponding author.}}
\address{\textsuperscript{1} School of Fundamental Science and Engineering, Waseda University, Tokyo, Japan\\
\textsuperscript{2} Waseda Research Institute for Science and Engineering, Waseda University, Tokyo, Japan\\
\textsuperscript{3} JST, PRESTO, 4-1-8 Honcho, Kawaguchi, Saitama, Japan
}
\begin{document}
%
\maketitle
\begin{abstract}
Recent state-of-the-art Learned Image Compression methods feature spatial context models, achieving great rate-distortion improvements over hyperprior methods. However, the autoregressive context model requires serial decoding, limiting runtime performance. The Checkerboard context model allows parallel decoding at a cost of reduced RD performance. We present a series of multistage spatial context models allowing both fast decoding and better RD performance. We split the latent space into square patches and decode serially within each patch while different patches are decoded in parallel. The proposed method features a comparable decoding speed to Checkerboard while reaching the RD performance of Autoregressive and even also outperforming Autoregressive. Inside each patch, the decoding order must be carefully decided as a bad order negatively impacts performance; therefore, we also propose a decoding order optimization algorithm. 
\end{abstract}
\begin{keywords}
Learned Image Compression, Spatial Context Model
\end{keywords}
\section{Introduction}
\label{sec:intro}

Learned Image Compression (LIC) features state-of-the-art rate-distortion (RD) performance with many recent methods outperforming the best hand-crafted approaches such as BPG \cite{bpg} and VVC \cite{vvc}. Many recent state-of-the-art LIC codecs utilize spatial context models \cite{NEURIPS2018_53edebc5,Cheng_2020_CVPR,Choi_2019_ICCV,koyuncu2022contextformer,He_2021_CVPR,TinyLIC,ccn,qian2022entroformer}, which infers the probability distribution of the current latent code based on already-decoded surrounding latent codes, greatly outperforming hyperprior-based methods \cite{balle2018variational}.

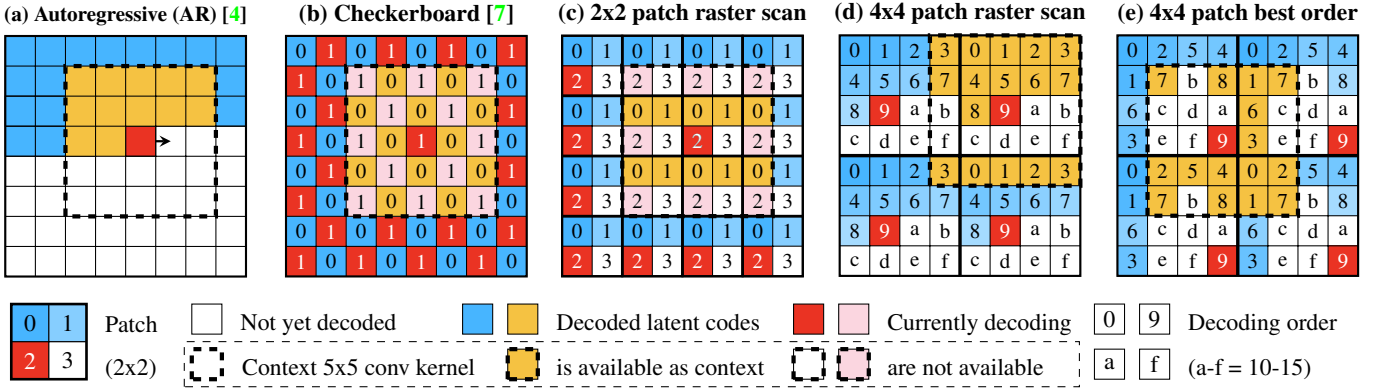
\begin{figure*}[btp]
\centerline{\tikzset{font={\fontsize{8pt}{12}\selectfont}}

\newcommand{\rulesep}{\unskip\ \vrule\ }

\begin{tikzpicture}[x=0.4cm,y=0.4cm,step=0.4cm]

\draw[fill=myblue] (0,5) rectangle (8,8) (0,4) rectangle (3,6);
\draw[fill=myred] (4,4) rectangle (5,5);
\fill[myyellow] (2,5) rectangle (7,7) (2,4) rectangle (4,5);
\draw (0,0) grid (8,8);
\draw[thick] (0,0) rectangle (8,8);
\draw[dashed, ultra thick] (2,2) rectangle (7,7);

\draw [draw=black,line width=0.3mm, -stealth] (5,4.5) -- (5.5,4.5);

\node[above,font=\bfseries\fontsize{8pt}{10}\selectfont] at (current bounding box.north) {(a) Autoregressive (AR) \cite{Cheng_2020_CVPR}};

\end{tikzpicture}\hspace{3mm}
\begin{tikzpicture}[x=0.4cm,y=0.4cm,step=0.4cm]

\foreach \y in {0,2,...,6}{
    \foreach \x in {0,2,...,6}{
        \fill[myblue] (\x,1+\y) rectangle (1+\x,2+\y) (1+\x,\y) rectangle (2+\x,1+\y);
        \fill[myred] (\x,\y) rectangle (1+\x,1+\y) (1+\x,1+\y) rectangle (2+\x,2+\y);
    }
}

\fill[myotherred!20] (2,2) rectangle +(5,5);
\fill[myyellow] (3,2) rectangle (4,3) (5,2) rectangle (6,3);
\fill[myyellow] (3,4) rectangle (4,5) (5,4) rectangle (6,5);
\fill[myyellow] (3,6) rectangle (4,7) (5,6) rectangle (6,7);
\fill[myyellow] (2,3) rectangle (3,4) (4,3) rectangle (5,4) (6,3) rectangle (7,4);
\fill[myyellow] (2,5) rectangle (3,6) (4,5) rectangle (5,6) (6,5) rectangle (7,6);

\foreach \y in {0,2,...,6}{
    \foreach \x in {0,2,...,6}{
        \node[white] at (\x + 0.5, \y + 0.5) {1};
        \node[white] at (\x + 1.5, \y + 1.5) {1};
        \node at (\x + 0.5, \y + 1.5) {0};
        \node at (\x + 1.5, \y + 0.5) {0};
    }
}
\foreach \y in {2,4,6}{
    \foreach \x in {2,4,6}{
        \node at (\x + 0.5, \y + 0.5) {1};
    }
}
\foreach \y in {3,5}{
    \foreach \x in {3,5}{
        \node at (\x + 0.5, \y + 0.5) {1};
    }
}

\fill[myred] (4, 4) rectangle (5,5);
\node[white] at (4.5, 4.5) {1};

\draw (0,0) grid (8,8);
\draw[dashed, ultra thick] (2,2) rectangle (7,7);
\draw[thick] (0,0) rectangle (8,8);

\node[above,font=\bfseries\fontsize{9pt}{10}\selectfont] at (current bounding box.north){(b) Checkerboard \cite{He_2021_CVPR}};

\end{tikzpicture}\hspace{2mm}
\begin{tikzpicture}[x=0.4cm,y=0.4cm,step=0.4cm]

\foreach \y in {0,2,...,6}{
    \foreach \x in {0,2,...,6}{
        \fill[myblue] (\x, \y+1) rectangle (\x+1, \y+2);
        \fill[myblue!66] (\x+1, \y+1) rectangle (\x+2, \y+2);
        \fill[myred] (\x, \y) rectangle (\x+1, \y+1);
        \fill[white] (\x+1, \y) rectangle (\x+2, \y+1);
    }
}

\fill[white] (2,2) rectangle (7,7);
\foreach \y in {2,4,6}{
    \foreach \x in {2,4,6}{
        \fill[myotherred!20] (\x,\y) rectangle +(1,1);
    }
}
\fill[myred] (4,4) rectangle (5,5);
\fill[myyellow] (2,5) rectangle (7,6) (2,3) rectangle (7,4);

\foreach \y in {0,2,...,6}{
    \foreach \x in {0,2,...,6}{
        \node at (\x + 0.5, \y + 1.5) {0};
        \node at (\x + 1.5, \y + 1.5) {1};
        \node[white] at (\x + 0.5, \y + 0.5) {2};
        \node at (\x + 1.5, \y + 0.5) {3};
        \draw[thick] (\x, \y) rectangle (\x + 2, \y + 2);
    }
}

\foreach \y in {2,4,6}{
    \foreach \x in {2,4,6}{
        \node at (\x + 0.5, \y + 0.5) {2};
    }
}

\node[white] at (4.5,4.5) {2};

\draw (0,0) grid (8,8);
\draw[line width=0.35mm] (2,0) -- (2,8) (4,0) -- (4,8) (6,0) -- (6,8) (0,2) -- (8,2) (0,4) -- (8,4) (0,6) -- (8,6);
\draw[dashed, ultra thick] (2,2) rectangle (7,7);
\draw[thick] (0,0) rectangle (8,8);

\node[above,font=\bfseries\fontsize{9pt}{10}\selectfont] at ($(current bounding box.north)$ ){(c) 2x2 patch raster scan};
\end{tikzpicture}\hspace{0.5mm}
\begin{tikzpicture}[x=0.4cm,y=0.4cm,step=0.4cm]

\foreach \y in {0,4}{
    \foreach \x in {0,4}{
        \fill[myblue] (\x, \y+3) rectangle +(1,1);
        \fill[myblue!94] (\x+1, \y+3) rectangle +(1,1);
        \fill[myblue!88] (\x+2, \y+3) rectangle +(1,1);
        \fill[myblue!82] (\x+3, \y+3) rectangle +(1,1);
        \fill[myblue!76] (\x, \y+2) rectangle +(1,1);
        \fill[myblue!70] (\x+1, \y+2) rectangle +(1,1);
        \fill[myblue!63] (\x+2, \y+2) rectangle +(1,1);
        \fill[myblue!56] (\x+3, \y+2) rectangle +(1,1);
        \fill[myblue!49] (\x, \y+1) rectangle +(1,1);
        \fill[myred] (\x+1,\y+1) rectangle +(1,1);
    }
}

\fill[white] (3,3) rectangle (8,8);
\fill[myred] (5,5) rectangle (6,6);
\fill[myyellow] (3,3) rectangle (8,4) (3,6) rectangle (8,8) (4,5) rectangle (5,6);

\foreach \y in {0,4}{
    \foreach \x in {0,4}{

        \node at (\x+0.5, \y+3.5) {0};
        \node at (\x+1.5, \y+3.5) {1};
        \node at (\x+2.5, \y+3.5) {2};
        \node at (\x+3.5, \y+3.5) {3};
        \node at (\x+0.5, \y+2.5) {4};
        \node at (\x+1.5, \y+2.5) {5};
        \node at (\x+2.5, \y+2.5) {6};
        \node at (\x+3.5, \y+2.5) {7};
        \node at (\x+0.5, \y+1.5) {8};
        \node[white] at (\x+1.5, \y+1.5) {9};
        \node at (\x+2.5, \y+1.5) {a};
        \node at (\x+3.5, \y+1.5) {b};
        \node at (\x+0.5, \y+0.5) {c};
        \node at (\x+1.5, \y+0.5) {d};
        \node at (\x+2.5, \y+0.5) {e};
        \node at (\x+3.5, \y+0.5) {f};
    }
}

\draw (0,0) grid (8,8);
\foreach \y in {0,4}{
    \foreach \x in {0,4}{
    }
}

\draw[line width=0.45mm] (4,0) -- (4,8) (0,4) -- (8,4);
\draw[thick] (0,0) rectangle (8,8);
\draw[dashed, ultra thick] (3,3) rectangle (8,8);

\node[above,font=\bfseries\fontsize{9pt}{10}\selectfont] at (current bounding box.north) {(d) 4x4 patch raster scan};
\end{tikzpicture}\hspace{1mm}
\begin{tikzpicture}[x=0.4cm,y=0.4cm,step=0.4cm]

\foreach \y in {0,4}{
    \foreach \x in {0,4}{
        \fill[myblue] (\x, \y+3) rectangle (\x+1, \y+4);
        \fill[myblue!94] (\x, \y+2) rectangle (\x+1, \y+3);
        \fill[myblue!88] (\x+1, \y+3) rectangle (\x+2, \y+4);
        \fill[myblue!82] (\x, \y) rectangle (\x+1, \y+1);
        \fill[myblue!76] (\x+3, \y+3) rectangle (\x+4, \y+4);
        \fill[myblue!70] (\x+2, \y+3) rectangle (\x+3, \y+4);
        \fill[myblue!63] (\x, \y+1) rectangle (\x+1, \y+2);
        \fill[myblue!56] (\x+1, \y+2) rectangle (\x+2, \y+3);
        \fill[myblue!49] (\x+3, \y+2) rectangle (\x+4, \y+3);
        \fill[myblue!42] (\x+3, \y) rectangle (\x+4, \y+1);
        \fill[white] (\x+2, \y) rectangle (\x+3, \y+1);
    }
}

\fill[myyellow] (1,2) rectangle (6,7);
\fill[white] (2,2) rectangle (3,3) (5,4) rectangle (6,6) (1,4) rectangle (3,6) (2,6) rectangle (3,7) (3,5) rectangle (4,6);

\foreach \y in {0,4}{
    \foreach \x in {0,4}{
        \node at (\x + 0.5, \y + 3.5) {0};
        \node at (\x + 0.5, \y + 2.5) {1};
        \node at (\x + 1.5, \y + 3.5) {2};
        \node at (\x + 0.5, \y + 0.5) {3};
        \node at (\x + 3.5, \y + 3.5) {4};
        \node at (\x + 2.5, \y + 3.5) {5};
        \node at (\x + 0.5, \y + 1.5) {6};
        \node at (\x + 1.5, \y + 2.5) {7};
        \node at (\x + 3.5, \y + 2.5) {8};
        \fill[myred] (\x+3,\y) rectangle (\x+4,\y+1);
        \node[white] at (\x + 3.5, \y + 0.5) {9};
        \node at (\x + 3.5, \y + 1.5) {a};
        \node at (\x + 2.5, \y + 2.5) {b};
        \node at (\x + 1.5, \y + 1.5) {c};
        \node at (\x + 2.5, \y + 1.5) {d};
        \node at (\x + 1.5, \y + 0.5) {e};
        \node at (\x + 2.5, \y + 0.5) {f};
    }
}

\draw[line width=0.45mm] (4,0) -- (4,8) (0,4) -- (8,4);
\draw (0,0) grid (8,8);
\draw[dashed, ultra thick] (1,2) rectangle (6,7);
\draw[thick] (0,0) rectangle (8,8);

\node[above,font=\bfseries\fontsize{9pt}{10}\selectfont] at (current bounding box.north) {(e) 4x4 patch best order};
\end{tikzpicture}}
\vspace{2mm}
\centerline{\tikzset{font={\fontsize{9pt}{12}\selectfont}}

\begin{tikzpicture}[x=0.4cm,y=0.4cm,step=0.4cm]
\draw[fill=white] (3,0) rectangle +(1,1) node (white0) {};
\node[right=0 of white0.east, anchor=north west] {Not yet decoded};

\draw[fill=myblue] (12,0) rectangle +(1,1) node (blue) {};
\draw[fill=myyellow] (13.5,0) rectangle +(1,1) node (yellow0) {};
\node[right=0 of yellow0.east, anchor=north west] {Decoded latent codes};

\draw[fill=myred] (23,0) rectangle +(1,1) node (red) {};
\draw[fill=myotherred!20] (24.5,0) rectangle +(1,1) node (otherred0) {};
\node[right=0 of otherred0.east, anchor=north west] {Currently decoding};

\draw[dashed,ultra thick] (3, -1.5) rectangle +(1,1) node (dashed) {};
\node[right=0 of dashed.east, anchor=north west] {Context 5x5 conv kernel};

\draw[fill=myyellow] (13.5,-1.5) rectangle +(1,1) node (yellow1) {};
\draw[dashed,ultra thick] (13.5,-1.5) rectangle +(1,1);
\node[right=0 of yellow1.east, anchor=north west] {is available as context};

\draw[fill=white] (23,-1.5) rectangle +(1,1) node (white1) {};
\draw[fill=myotherred!20] (24.5,-1.5) rectangle +(1,1) node (otherred1) {};
\draw[dashed,ultra thick] (23,-1.5) rectangle +(1,1);
\draw[dashed,ultra thick] (24.5,-1.5) rectangle +(1,1);
\node[right=0 of otherred1.east, anchor=north west] {are not available};

\draw[dashed] (2.75, -1.75) rectangle (32.5, -0.25);

\draw (33,0) rectangle +(1,1);
\draw (34.5,0) rectangle +(1,1) node(num9) {};
\draw (33,-1.5) rectangle +(1,1);
\draw (34.5,-1.5) rectangle +(1,1) node (numf) {};
\node at (33.5,0.5) {0};
\node at (35,0.5) {9};
\node at (33.5,-1) {a};
\node at (35,-1) {f};
\node[right=0 of num9.east, anchor=north west] (dec_order) {Decoding order};
\node[right=0 of numf.east, anchor=north west] (a_f) {(a-f = 10-15)};

\draw[fill=myred] (-3, -1.5) rectangle +(1.25, 1.25) node[white,pos=.5] {2};
\draw[fill=white] (-1.75, -1.5) rectangle +(1.25, 1.25) node[pos=.5] (patch3) {3};
\draw[fill=myblue] (-3, -0.25) rectangle +(1.25, 1.25) node[pos=.5] {0};
\draw[fill=myblue!66] (-1.75, -0.25) rectangle +(1.25, 1.25) node[pos=.5] (patch1) {1};
\draw[line width=0.35mm] (-3, -1.5) rectangle +(2.5, 2.5);
\node[left=36 of dec_order.west, anchor=west] {Patch};
\node[left=36 of a_f.west, anchor=west] {(2x2)};
\end{tikzpicture}}
\caption{Illustration of different context models; (c) (d) (e) are examples of proposed context models.}
\label{f:related_work}
\end{figure*}

Many of the previous spatial context models have an autoregressive structure: as shown in Figure \ref{f:related_work} (a), latent codes are decoded in raster scan order. Its use can be widely seen in previous state-of-the-art works such as \cite{NEURIPS2018_53edebc5,Cheng_2020_CVPR,Choi_2019_ICCV}. However, as the decoding of later latent codes always rely on the information from previous ones, these models only run serially and have slow decoding speeds.  

The Checkerboard context model \cite{He_2021_CVPR}, shown in Figure \ref{f:related_work} (b), speeds up this process by decoding half of the latent codes using only hyperprior. Then, the remaining half is decoded using both hyperprior and already-decoded latent codes as context. It enables parallel execution and features fast decoding speed at a cost of reduced rate-distortion performance. Is it possible to have good RD performance and still be fast?

We show that it is possible, by presenting a series of spatial context models with varying numbers of decoding stages. Our context models split the image latent space into square patches, then decode autoregressively only within each patch. Since latent codes in different patches do not rely on each other, different patches can be decoded in parallel. 

The decoding order within each patch must be carefully decided, as we show that a bad decoding order can damage performance. We propose an optimization algorithm to find the best decoding order by estimating the effects of different orders on the Random Mask model \cite{He_2021_CVPR}. We finally show that our context models, with optimized decoding orders, outperform Checkerboard and reach the RD performance of the autoregressive context model and even also outperform it.

\section{Related Work}
\label{sec:related}

Despite Checkerboard \cite{He_2021_CVPR}, most other attempts to address the serial execution bottleneck of spatial context models still revolve around the autoregressive structure \cite{koyuncu2022contextformer,ccn}. For example, CCN \cite{ccn} sees the image latent space as a 3D cube and decodes all latent on the same plane in one stage; however, the number of decoding stages still depends on the image size.

Another set of methods features a constant number of decode stages, including our proposed method. There are models featuring the Checkerboard pattern \cite{qian2022entroformer,He_2022_CVPR}. One particular work having some similarities with our proposed method features an extension to Checkerboard \cite{TinyLIC}. It splits the anchors and non-anchors further into two halves and decodes the latent codes in a diagonal order. However, its performance is limited by the decoding order, as discussed later in Section \ref{ssec:decoding_order} and \ref{ssec:ablation} and shown in Figure \ref{f:random_mask} (b).

Another class of context models exploits channel-wise information \cite{koyuncu2022contextformer,He_2022_CVPR,9190935}, which has also been gaining attention recently and used in state-of-the-art architectures, although this is beyond the scope of this paper as we will focus on discussing spatial context models.

\section{Proposed Method}
\label{sec:method}

The decoding of an image in LIC with a spatial context model \cite{NEURIPS2018_53edebc5,Cheng_2020_CVPR} consists of: (1) the entropy decoding and synthesis of hyperprior; (2) the retrieval of spatial context information (if any) through a convolution layer, and inference of latent probability distribution from concatenated hyperprior and context, and the entropy decoding of latent codes; (3) the image synthesis from the latent space. We focus the discussion on (2), a recurrent process potentially bottlenecking performance. We refer to a context model as $m$-stage if (2) is repeated $m$ times. We base our discussion on \cite{Cheng_2020_CVPR} although our method is not limited to this architecture. As we focus on discussing different context models, we do not modify other model structures and use a 5x5 convolution kernel for extracting context.

\subsection{The multistage context models}
\label{ssec:multistage}

Regarding rate-distortion performance, the main limitation of Checkerboard is the high proportion of anchor latent codes\footnote{Those that are encoded/decoded with hyperprior only.}. Unable to benefit from spatial context, anchor latent codes will generally require more bits to encode. As shown in Figure \ref{f:related_work} (b), Checkerboard can be seen as a 2-stage context model; 50\% of latent codes are decoded in stage 0 as anchors, to serve as spatial context information for stage 1. This means half of the latent representation cannot benefit from the bit-rate reduction provided by the context model.

It is therefore intuitive that decreasing the proportion of anchors boosts RD performance. We achieve this with more decoding stages. We split the latent space into equal squares of $n \times n$ patches, and decode it in $n^2$ stages. In each stage, we decode one latent code\footnote{In reality, it is not just one, but latent codes from all channels in the same spatial position. Unless necessary, we ignore the channel dimension throughout this paper for simplicity as we focus on the two spatial dimensions.} from each patch; that is, $\frac{hw}{n^2}$ latent codes are decoded in each stage. As two examples, we divide the latent space into $2 \times 2$ patches, shown in Figure \ref{f:related_work} (c), in which the decoding happens in 4 stages; and $4 \times 4$ patches, shown in Figure \ref{f:related_work} (d) and (e) and decoded in 16 stages. 

The decoding of latent codes is performed in the order shown by ascending hexadecimal numbers. For example, (c) and (d) show two raster scan orders. Only the latent codes decoded by previous stages are available and used as context information through a 5x5 convolution. For instance, in (c), all latent codes of stage 2 of each patch are being decoded in parallel; in the 5x5 receptive fields, latent codes of previous stages 0 and 1 can be used as context; in contrast, those of stages 2 and 3 cannot since they have not been decoded. 

This way, only the latent codes at stage 0 must be decoded without any context information. Our multistage context models reduce the proportion of anchors to $\frac{1}{n^2} \times 100\%$, 25\% in the case of a $2 \times 2$ model and 6.25\% in $4 \times 4$. 

\subsection{Decoding order optimization algorithm}
\label{ssec:search}

\begin{figure}[btp]
\centerline{\begin{tikzpicture}[x=0.24cm,y=0.24cm,step=0.24cm]

\foreach \y in {0,2,...,6}{
    \foreach \x in {0,2,...,6}{
        \fill[myred] (\x,\y+1) rectangle +(1,1);
    }
}
\fill[white] (2,3) rectangle +(5,5);
\foreach \y in {3,5,7}{
    \foreach \x in {2,4,6}{
        \fill[myotherred!20] (\x,\y) rectangle +(1,1);
    }
}
\fill[myred] (4,5) rectangle +(1,1);
\draw[dashed,ultra thick] (2,3) rectangle +(5,5);

\foreach \y in {0,2,...,6}{
    \foreach \x in {0,2,...,6}{
        \fill[myblue] (9+\x,\y+1) rectangle +(1,1);
        \fill[myred] (9+\x+1,\y+1) rectangle +(1,1);
    }
}
\fill[white] (9+3,3) rectangle +(5,5);
\foreach \y in {3,5,7}{
    \foreach \x in {3,5,7}{
        \fill[myotherred!20] (9+\x,\y) rectangle +(1,1);
    }
}
\fill[myred] (9+5,5) rectangle +(1,1);
\foreach \y in {3,5,7}{
    \foreach \x in {4,6}{
        \fill[myyellow] (9+\x,\y) rectangle +(1,1);
    }
}
\draw[dashed,ultra thick] (9+3,3) rectangle +(5,5);

\foreach \y in {0,2,...,6}{
    \foreach \x in {0,2,...,6}{
        \fill[myblue] (18+\x,\y+1) rectangle +(1,1);
        \fill[myblue!66] (18+\x+1,\y+1) rectangle +(1,1);
        \fill[myred] (18+\x,\y) rectangle +(1,1);
    }
}
\fill[white] (18+2,2) rectangle +(5,5);
\foreach \y in {2,4,6}{
    \foreach \x in {2,4,6}{
        \fill[myotherred!20] (18+\x,\y) rectangle +(1,1);
    }
}
\fill[myred] (18+4,4) rectangle +(1,1);
\fill[myyellow] (18+2,3) rectangle +(5,1) (18+2,5) rectangle +(5,1);
\draw[dashed,ultra thick] (18+2,2) rectangle +(5,5);

\foreach \graphstart in {0, 9, 18, 27}{
    \foreach \y in {0,2,...,6}{
        \foreach \x in {0,2,...,6}{
            \draw[thick] (\graphstart+\x,\y) rectangle (\graphstart+\x+2,\y+2);
        }
    }
    \draw (\graphstart,0) grid +(8,8);
}

\foreach \y in {0,2,...,6}{
    \foreach \x in {0,2,...,6}{
        \fill[myblue] (27+\x,\y+1) rectangle +(1,1);
        \fill[myblue!66] (27+\x+1,\y+1) rectangle +(1,1);
        \fill[myblue!33] (27+\x,\y) rectangle +(1,1);
        \fill[myred] (27+\x+1,\y) rectangle +(1,1);
    }
}
\fill[white] (27+3,2) rectangle +(5,5);
\foreach \y in {2,4,6}{
    \foreach \x in {3,5,7}{
        \fill[myotherred!20] (27+\x,\y) rectangle +(1,1);
    }
}
\fill[myred] (27+5,4) rectangle +(1,1);
\fill[myyellow] (27+3,3) rectangle +(5,1) (27+3,5) rectangle +(5,1);
\fill[myyellow] (27+4,2) rectangle +(1,5) (27+6,2) rectangle +(1,5);
\draw[dashed,ultra thick] (27+3,2) rectangle +(5,5);

\foreach \graphstart in {0, 9, 18, 27}{
    \foreach \y in {0,2,...,6}{
        \foreach \x in {0,2,...,6}{
            \draw[thick] (\graphstart+\x,\y) rectangle (\graphstart+\x+2,\y+2);
        }
    }
    \draw (\graphstart,0) grid +(8,8);
}

\end{tikzpicture}}
\caption{Illustration of decoding stages of $2 \times 2$ patch raster scan with 5x5 convolution kernel.}
\label{f:2x2}
\end{figure}
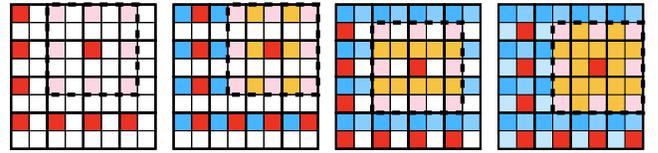

There are many possible decoding orders in our proposed multistage context models. Figure \ref{f:related_work} (d) and (e) show two different possible orders in $4 \times 4$ patches. In different decoding orders, the latent codes available as context information also vary. For example, Figure \ref{f:2x2} shows the decoding stages and corresponding available context positions of the decoding order shown in Figure \ref{f:related_work} (c); other decoding orders may not have access to those same context positions. This results in some decoding orders performing better than others as they can better utilize context information.

However, training models in all arrangements of decoding orders and evaluating them for the best is impossible, especially with more stages. For our $4 \times 4$ model, or 16-stages, there are 16! or 20+ trillion possible arrangements; it is impossible to train this many models even with supercomputers.

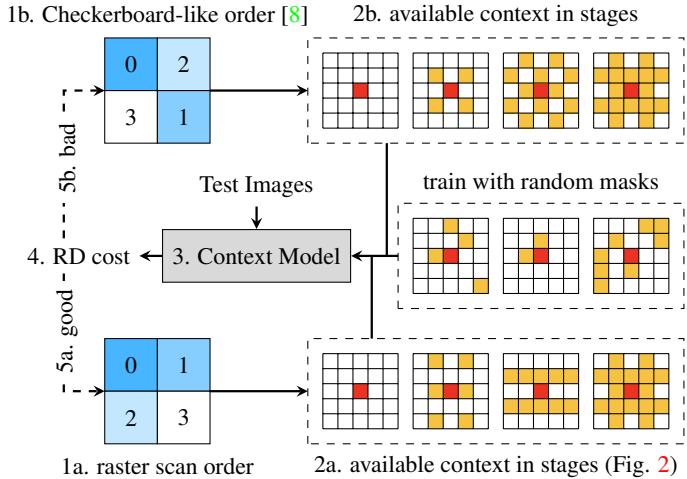
\begin{figure}[btp]
\centerline{\tikzset{font={\fontsize{9pt}{12}\selectfont}}

\begin{tikzpicture}[x=0.2cm,y=0.2cm,step=0.2cm]
\draw[dashed] (0,0) rectangle (19,7);
\node at (0, 3.5) (outline_anchor) {};
\fill[myred] (3,3) rectangle (4,4) (9,3) rectangle (10,4) (15,3) rectangle (16,4);
\fill[myyellow] (2,3) rectangle (3,4) (3,5) rectangle (4,6) (4,4) rectangle (5,5) (5,1) rectangle (6,2);
\fill[myyellow] (8,3) rectangle (9,4) (9,4) rectangle (10,5) (4,4) rectangle (5,5) (5,1) rectangle (6,2);
\fill[myyellow] (13,1) rectangle (14,4) (14,4) rectangle (15,5) (15,2) rectangle (16,3) (17,4) rectangle (18,6) (16,5) rectangle (17,6);
\draw (1,1) grid (6,6) (7,1) grid (12,6) (13,1) grid (18,6);
\node at (9.5, 8.5) {train with random masks};

\node[draw,rectangle, fill=gray!30, minimum height=0.7cm, minimum width=2.5cm, anchor=east, left=2.5 of outline_anchor] (random_mask) {3. Context Model};
\draw[line width=0.3mm, -stealth] (0, 3.5) -- (random_mask);

\node[above=1.5 of random_mask.north] (images) {Test Images};
\draw[line width=0.3mm, -stealth] (images.south) -- (random_mask.north);

\node[left=1.5 of random_mask.west, anchor=east] (rd_cost) {4. RD cost};
\draw[line width=0.3mm, -stealth] (random_mask.west) -- (rd_cost.east);

\draw[dashed] (-6,-9) rectangle (19,-2);
\node at (6.5, -10.5) {2a. available context in stages (Fig. \ref{f:2x2})};
\draw[line width=0.3mm] (-1.75, -2) -- (-1.75, 3.5);
\fill[myred] (-3,-6) rectangle (-2,-5) (3,-6) rectangle (4,-5) (9,-6) rectangle (10,-5) (15,-6) rectangle (16,-5);
\fill[myyellow] (2,-8) rectangle (3,-7) (2,-6) rectangle (3,-5) (2,-4) rectangle (3,-3) (4,-8) rectangle (5,-7) (4,-6) rectangle (5,-5) (4,-4) rectangle (5,-3); 
\fill[myyellow] (7,-7) rectangle (12,-6) (7,-5) rectangle (12,-4);
\fill[myyellow] (13,-7) rectangle (18,-6) (13,-5) rectangle (18,-4) (14,-8) rectangle (15,-3) (16,-8) rectangle (17,-3);
\draw (-5,-8) grid (0,-3) (1,-8) grid (6,-3) (7,-8) grid (12,-3) (13,-8) grid (18,-3);

\draw[fill=myblue] (-19.5, -5.5) rectangle (-16, -2) node[pos=.5] {0};
\draw[fill=myblue!66] (-16, -5.5) rectangle (-12.5, -2) node[pos=.5] {1};
\draw[fill=myblue!33] (-19.5, -9) rectangle (-16, -5.5) node[pos=.5] {2};
\draw[fill=white] (-16, -9) rectangle (-12.5, -5.5) node[pos=.5]{3};
\node at (-16, -10.5) {1a. raster scan order};

\draw[line width=0.3mm, -stealth] (-12.5, -5.5) -- (-6, -5.5);

\draw[dashed] (-6, 11) rectangle +(25,7);
\node at (6.5, 19.5) {2b. available context in stages};
\draw[line width=0.3mm] (-0.75, 11) -- (-0.75, 3.5);

\draw[fill=white] (-19.5, 11) rectangle +(3.5, 3.5) node[pos=.5] {3};
\draw[fill=myblue!66] (-16, 11) rectangle +(3.5, 3.5) node[pos=.5] {1};
\draw[fill=myblue] (-19.5, 14.5) rectangle +(3.5, 3.5) node[pos=.5] {0};
\draw[fill=myblue!33] (-16, 14.5) rectangle +(3.5, 3.5) node[pos=.5]{2};
\node at (-16, 19.5) {1b. Checkerboard-like order \cite{TinyLIC}};

\fill[myred] (-3,14) rectangle +(1,1) (3,14) rectangle +(1,1) (9,14) rectangle +(1,1) (15,14) rectangle +(1,1);
\fill[myyellow] (2,13) rectangle +(1,1) (2,15) rectangle +(1,1) (4,13) rectangle +(1,1) (4,15) rectangle +(1,1);
\foreach \x in {7,9,11} {
    \foreach \y in {13,15} {
        \fill[myyellow] (\x,\y) rectangle +(1,1);
    }
}
\foreach \x in {8,10} {
    \foreach \y in {12,14,16} {
        \fill[myyellow] (\x,\y) rectangle +(1,1);
    }
}
\fill[myyellow] (13,13) rectangle (18,14) (13,15) rectangle (18,16) (14,12) rectangle (15,17) (16,12) rectangle (17,17);

\draw (-5,12) grid +(5,5) (1,12) grid +(5,5) (7,12) grid +(5,5) (13,12) grid +(5,5);

\draw[line width=0.3mm, -stealth] (-12.5, 14.5) -- (-6, 14.5);

\draw[line width=0.3mm, -stealth, dashed] ($ (rd_cost.north) +(-1,0) $) |- (-19.5, 14.5);
\node[rotate=90,fill=white] at(-22, 10) {5b. bad};
\draw[line width=0.3mm, -stealth, dashed] ($ (rd_cost.south) +(-1,0) $) |- (-19.5, -5.5);
\node[rotate=90,fill=white] at(-22, -1.5) {5a. good};
\end{tikzpicture}}
\caption{Evaluation of different $2 \times 2$ decoding orders.}
\label{f:random_mask}
\end{figure}

Therefore, we propose an estimation algorithm simulating the effect of different decoding orders. We reproduce the Random Mask model introduced in \cite{He_2021_CVPR}, which follows the typical structure of spatial context models but allows an extra input to temporarily zero kernel weights in context convolution. As Figure \ref{f:random_mask} shows, in each training step, a 5x5 mask is randomly generated, and the weights under non-masked positions in the context convolution kernel are backed up and zeroed so that the model is only allowed to utilize masked latent code positions as context information. The zeroed weights are restored after each step. During backward propagation, only the weights under masked positions are updated. The context model gradually learns to produce good enough inferences for any set of given latent code positions in the 5x5 kernel.  

Our approach to evaluating a decoding order is shown in Figure \ref{f:random_mask}. (1) We enumerate all arrangements of decoding orders in a $n \times n$ patch with a depth-first search. (2) For each decoding order, we find which context positions are available to be used as context in each stage by the process shown in Figure \ref{f:2x2}. We always use all available positions in the 5x5 convolution kernel. (3) We feed the position mask into the Random Mask model, to limit it to only use these positions as context. (4) The Random Mask model evaluates the RD costs by encoding test images with context from these masked positions. (5) We estimate the performance of each decoding pattern by comparing the RD costs; lower is better. 

Here we choose a 5x5 convolution kernel because to learn from and search larger masks such as 7x7 is too computationally heavy; 5x5 is a sweet spot \cite{Cheng_2020_CVPR, He_2021_CVPR}. It is also worth noting that the Random Mask model is only for experiments and not used in actual compression, as we optimize it to perform generally well with any mask. We then hard-code the best decoding order found into the codec, and train models optimizing only for it, to achieve the best RD performance.

\subsection{Practical patch sizes}
\label{ssec:patchsize}

A typical synthesis transformation downscale an image of $3 \times H \times W$ to $M \times \frac{H}{16} \times \frac{W}{16}$. To split such latent space into $n \times n$ patches, $H$ and $W$ must be padded to multiples of $16n$. A typical hyperprior transformation downscale the latent space further down to $N \times \frac{H}{64} \times \frac{W}{64}$, so it is already required that $H$ and $W$ padded to multiples of $64$ in hyperprior-based LICs.

With the addition of our context model, the image must be padded to a size that is a multiple of $\lcm(16n, 64)$. This poses a problem when $n = 3$, for instance, as $\lcm(48, 64)=192$ requires a large amount of padding for many images, making such patch sizes impractical in real use. Patch sizes of $2 \times 2$ and $4 \times 4$ are not affected by this as $\lcm(16n, 64)$ of both sizes are perfectly 64 so that no extra padding is needed.

\section{Experiments}
\label{sec:experiments}

We experiment with two variations of our context models: $2 \times 2$ and $4 \times 4$. As previously explained, the $3 \times 3$ variation was trained but failed to produce outstanding results due to its impracticality; therefore, we skip it for detailed analysis. 

We base our implementation on CompressAI's Cheng2020-Attention \cite{compressai} but change it to use GMM with the number of normal distributions $K=3$ to follow the original settings and swap out the autoregressive context model with the proposed.

\subsection{Best decoding orders}
\label{ssec:decoding_order}

We use the method described in Section \ref{ssec:search} to search for the best decoding orders in both $2 \times 2$ and $4 \times 4$ variation. We evaluate with a quality parameter $\lambda=0.0035$ variation of the Random Mask model and use a few random images as input. 

We find that for the $2 \times 2$ model, the raster scan order shown in Figure \ref{f:related_work} (c) is already the best decoding order. For $4 \times 4$, the raster scan order shown in Figure \ref{f:related_work} (d) is not optimal. One of the best decoding orders is shown in Figure \ref{f:related_work} (e). We use the best decoding orders for later experiments.

We noticed that good decoding orders always refer to four adjacencies in every stage. This can be justified by the original Random Mask experiment \cite{He_2021_CVPR}, which concluded that four adjacency produces the most bitrate reduction. This explains why the raster scan order in $2 \times 2$ is superior to the Checkerboard-like pattern \cite{TinyLIC}: the latter fails to refer to any four adjacencies in stage 1 as shown in Figure \ref{f:random_mask} (2b), so the stage cannot achieve good bitrate reduction. Referring to spatial context from different directions also benefits accuracy; as the $4 \times 4$ raster scan mostly refers to top and left latent codes, it is outperformed by the best order which exploits spatial dependencies from multiple directions.

\subsection{Rate-distortion performance}

We train our models on a subset of around 360,000 images randomly selected from the Open Image Dataset \cite{OpenImages}. We train 6 variations of $2 \times 2$ and $4 \times 4$ models, each with a quality parameter $\lambda \in \{0.0018,0.0035,0.0067,0.013,0.025,0.0483\}$ and optimize to target PSNR. For the first 3 $\lambda$ values, we use $N=M=128$, and $N=M=192$ for the rest, where $N$ is the number of channels in convolutions and $M$ is that in entropy bottleneck. We evaluate our models on Kodak Image Dataset \cite{kodak} and CLIC Professional \cite{CLIC2020}. We compare our two models with two baselines: (1) Autoregressive (AR) \cite{Cheng_2020_CVPR} in Fig. \ref{f:related_work} (a); (2) Checkerboard \cite{He_2021_CVPR} (on Kodak only) in Fig. \ref{f:related_work} (b). For the baselines, we use data from corresponding works. 

\begin{figure}[btp]
\centering
\includegraphics[width=\columnwidth]{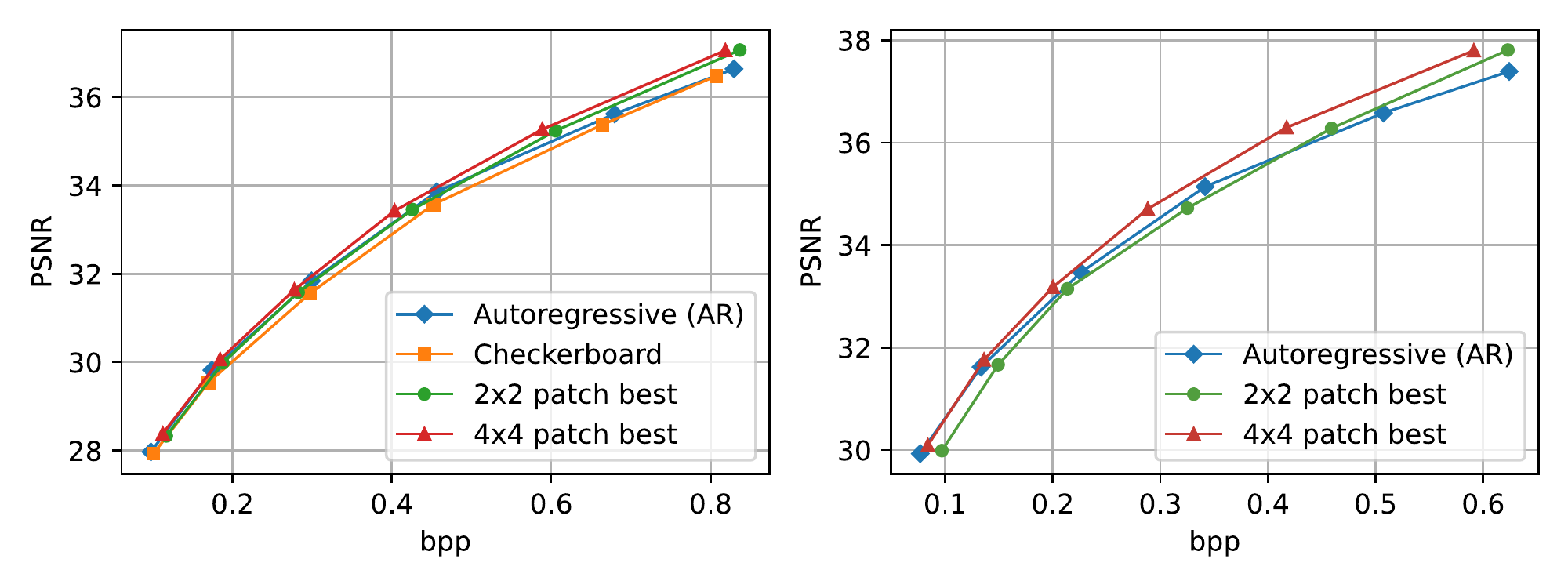}
\caption{Rate-distortion curves of the different context models on Kodak (Left) and CLIC Professional (Right).}
\label{f:psnr}
\end{figure}

Figure \ref{f:psnr} shows the RD curves with different context models. On Kodak, the proposed context models outperform Checkerboard, and both the $2 \times 2$ and $4 \times 4$ variations reach the RD performance of Autoregressive (AR). At higher bitrates, our context models outperform AR, and $4 \times 4$ shows around -2\% RD cost at all bitrates over $2 \times 2$. On CLIC, compared with AR, although $2 \times 2$ shows reduced performance at lower bitrates, $4 \times 4$ greatly outperforms AR and shows over -5\% RD cost at all bitrates over $2 \times 2$.

\subsection{Runtime performance}

\begin{table}[t]
\caption{Decoding time of context models on RTX 3080. The models evaluated have $N=M=128$. Evaluted on Kodak.}
\label{t:speed}
\vspace{1mm}
\centering
\begin{tabular}{|c|c||c|c|} 
 \hline
  Autoregressive (AR) & Checkerboard & \textbf{2x2} & \textbf{4x4} \\
 \hline\hline
 8574ms & 76ms & 85ms & 97ms \\
 \hline
\end{tabular}
\end{table}

We evaluate the decoding speed of the proposed context models and compare them against Autoregressive (AR) and Checkerboard. We use models with $N=M=128$ as they are more likely to be used in performance-critical situations and average the run time on Kodak. For a fair comparison, we use the same implementations except for necessary differences in the context model. As shown in Table \ref{t:speed}, our models are much faster than AR and are only slightly slower than Checkerboard due to an increased number of stages, which is an acceptable tradeoff due to the superior RD performance.

\subsection{Ablation study on decoding orders}
\label{ssec:ablation}

\begin{table}[t]
\caption{RD cost comparison of decoding orders.}
\label{t:ablation}
\begin{tabularx}{\columnwidth}{|c|>{\centering\arraybackslash}X||c|c||c|c|} 
 \hline
 \multicolumn{2}{|c||}{RD cost \%} & \multicolumn{2}{c||}{$\mathbf{\lambda}$\textbf{=0.0035}} & \multicolumn{2}{c|}{$\mathbf{\lambda}$\textbf{=0.025}} \\
 \hline
 \multicolumn{2}{|c||}{Dataset} & Kodak & CLIC & Kodak & CLIC \\
 \hline\hline
 \multirow{3}*{4x4} & best & \multicolumn{2}{c||}{0\%} & \multicolumn{2}{c|}{0\%} \\
 \cline{2-6}
 & raster & \textbf{+1.81\%} & \textbf{+2.97\%} & \textbf{+0.75\%} & \textbf{+1.17\%} \\
 \cline{2-6}
 & worst & \textbf{+2.09\%} & \textbf{+3.39\%} & \textbf{+0.89\%} & \textbf{+1.35\%} \\
 \hline\hline
 \multirow{2}*{2x2}& best & \multicolumn{2}{c||}{0\%} & \multicolumn{2}{c|}{0\%} \\
 \cline{2-6}
 & worst & \textbf{+2.49\%} & \textbf{+4.39\%} & \textbf{+0.82\%} & \textbf{+1.73\%} \\
 \hline
\end{tabularx}
\end{table}

We perform an ablation study to show the importance of good decoding order. To compare with the best decoding order, we train a set of extra models at $\lambda \in \{0.0035, 0.025\}$. For $4 \times 4$, we compare the best decoding order with the raster scan decoding order shown in Figure \ref{f:related_work} (d), and the least efficient (worst) decoding order. For $2 \times 2$, we compare the best decoding order (the raster scan order) shown in Figure \ref{f:related_work} (c), \ref{f:2x2} and \ref{f:random_mask} (a), with the Checkerboard-like order \cite{TinyLIC} shown in Figure \ref{f:random_mask} (b), as it was evaluated by our algorithm as the least efficient (worst). As shown in Table \ref{t:ablation}, increments in RD costs are seen with the less efficient decoding orders, especially at a lower bitrate. By evaluating the best decoding orders using the proposed algorithm before training context models correspondingly, we avoid leaving RD performance on the table.

\section{Conclusion}
\label{sec:conclusion}

In this work, we proposed splitting the latent space into even-sized square patches and decoding in an optimized order within patches. The proposed context models can have both RD performance comparable to and even outperforming the autoregressive context model and fast parallel decoding.

We showcased the performance of models with $2 \times 2$ and $4 \times 4$ patches. The rate-distortion performance is potentially further exploitable with larger patches. However, not only the padding problem described in Section \ref{ssec:patchsize} will become more dominant, but there is also a diminishing return that even a lot more stages bring no much better performance, as the extra context information is no longer efficiently utilized. We therefore consider $2 \times 2$ and $4 \times 4$ sweet spots. 

As future work, our work can be integrated with the more recent SOTA architectures with spatial context, or combined with channel-wise context \cite{9190935} to create better channel-spatial context models similar to SCCTX in ELIC \cite{He_2022_CVPR}, to achieve even better RD performance.

\section{Acknowledgement}
\label{sec:acknowledgement}

This work was supported in part by NICT No. 03801, JST PRESTO JPMJPR19M5, JSPS Grant 21K17770, Kenjiro Takayanagi Foundation, and Foundation of Ando laboratory.

\vfill\pagebreak

\bibliographystyle{IEEEtran}
\bibliography{icassp}

\end{document}